\documentclass{article}
\usepackage[final,sglblindworkshop]{neurips_2026}
\workshoptitle{GlobalSouthAI}
\usepackage[utf8]{inputenc}
\usepackage[T1]{fontenc}
\usepackage{longtable}
\usepackage{hyperref,url,booktabs,amsfonts,nicefrac,microtype,xcolor,multirow,amsmath,graphicx}

\title{Indic-Task Gains and Regressions\\after Reinforcement Learning}
\author{%
  \textbf{Sutanshu Saksena Raj} \quad \textbf{Daksh Maan} \quad \textbf{Avinash Benki} \\
  \textbf{Parijat Chatterjee} \quad \textbf{Pasunuti Prasanjith} \quad \textbf{Purusharth Malik} \\[2pt]
  \normalfont Gnani.AI \\[1pt]
  \normalfont\texttt{\{avinash, sutanshu.raj\} @ gnani.ai}%
}

\begin{document}
\maketitle

\begin{abstract}
We run reinforcement learning (RL) with GRPO across four environments, starting from a supervised fine-tuned 30B-A3B mixture-of-experts model covering English and eight Indic languages. We compare the model before and after RL across 34 benchmarks in one evaluation campaign. MATH improves by 36.07 points, and most Indic benchmarks also improve. Yet XQuAD-Indic F1 falls 19.32 points as English answers rise from 8.7\% to 91.7\%; IndicQA falls 8.64, while English XQuAD gains 11.71 F1. A second RL run from the same supervised initialization adds an Indic environment together with language and repetition rewards. Compared with the released model, it improves both passage-answering and instruction-following benchmarks, along with preserving MATH performance. But it loses ground on reasoning and coding benchmarks. Several training choices changed together, so we cannot isolate the effect of any one of the reward terms. Script compliance remains unresolved. Adding a language reward did not ensure use of the requested script. Passage answering and script compliance therefore need direct evaluation, \mbox{even when other benchmark scores improve.}
\end{abstract}

\section{Introduction}
Reinforcement learning after supervised fine-tuning trains a model using rewards assigned to its responses. For models serving underrepresented languages, the choice of rewards matters beyond performance on reasoning tasks. Users also need answers in the requested language and script~\citep{jayakumar-etal-2026-indicifeval}, and answers based on documents they provide~\citep{singh-etal-2024-indicgenbench}. A reward that scores mathematical correctness or valid tool calls does not necessarily check these requirements. The question is whether gains on the rewarded tasks come at the expense of \mbox{other abilities learned during supervised fine-tuning.}

We study this question in a 30B-A3B mixture-of-experts model covering English and eight Indic languages. We first compare a supervised checkpoint with its released RL descendant across 34 benchmarks. We then examine passage answering across English and Indic languages and test a separate RL run from the same supervised checkpoint with added Indic training and reward terms. This case study reports the gains and regressions from both runs, including changes in script compliance and answer formatting. Section~\ref{sec:setup} describes the checkpoints and training setup.

\subsection{Contributions}
\begin{enumerate}
\item A matched evaluation across 34 benchmarks that documents gains and regressions after RL, with a comparison of passage answering across English and Indic languages.
\item A comparison with a separate RL run from the same supervised checkpoint, showing how added Indic training changes task performance and script compliance, including the trade-offs on reasoning and coding benchmarks.
\end{enumerate}

\section{Related work}
Multilingual instruction tuning can transfer task behavior across languages~\citep{muennighoff-etal-2023-crosslingual,ustun-etal-2024-aya}. Studies also examine the choice between monolingual and multilingual training mixtures~\citep{chen-etal-2024-monolingual}. Indic resources emphasize native and generative evaluation rather than relying only on translated benchmarks~\citep{khan-etal-2024-indicllmsuite,singh-etal-2024-indicgenbench}. Work on language adaptation documents losses in previously learned behavior, especially with parameter-efficient methods~\citep{aggarwal-etal-2024-maple,toukmaji-flanigan-2025-lowresource,hu2021lora}. We examine changes after reinforcement learning rather than supervised adaptation. Closest to our setting, \citet{li-etal-2025-impact} find that RL induces language mixing in bilingual reasoning traces and that this can help performance. Our focus is the final answer, where using the requested language and script matters to the user.

\section{Training and evaluation}
\label{sec:setup}
\paragraph{Starting checkpoint.} The model has 30B total parameters and 3B active parameters, uses a mixture-of-experts architecture~\citep{nvidia2025nemotron}, and supports a context length of 131,072 tokens. We fine-tuned all parameters on a corpus containing 64.0\% English and 36.0\% Indic data across eight languages, in native and romanized script. We call the resulting checkpoint \textsc{SFT-Init}. It was exported partway through the final training pass. Appendix~\ref{app:sft} gives further training details.

\paragraph{Released RL model.} Starting from \textsc{SFT-Init}, we trained \textsc{RL-Base} with a GRPO variant~\citep{shao2024deepseekmath} using a leave-one-out baseline. It samples 16 completions per prompt across four environments: judged mathematics, JSON schema conformance, a tool-using workplace assistant, and multiple-choice question answering. Each environment provides prompts and scores the model's responses. There are no explicit language, script, grounding, length, or repetition reward terms. The run used 763 GPU-hours. The checkpoint is available at \url{https://huggingface.co/gnani/gnani-evon-v3.3-30B-A3B}. Dynamic sampling discards groups with identical completion scores, so input proportions need not match training proportions. Appendix~\ref{app:rl} describes the RL setup.

\begin{figure}[t]
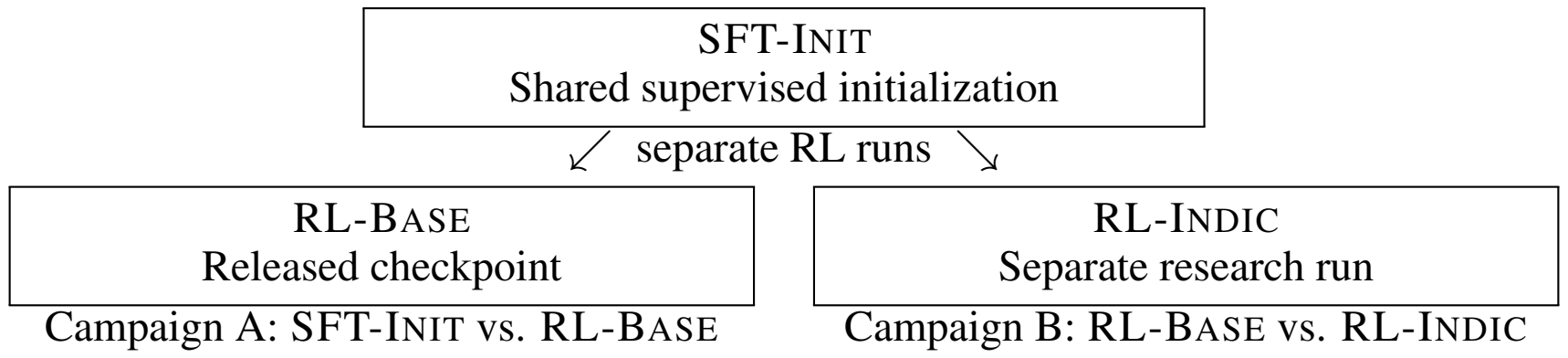

\centering\small
\begin{tabular}{@{}c@{}}
\fbox{\begin{minipage}{0.42\linewidth}\centering
\textsc{SFT-Init}\\Shared supervised initialization
\end{minipage}}\\[2pt]
$\swarrow$\; separate RL runs \;$\searrow$\\[2pt]
\begin{tabular}{@{}cc@{}}
\fbox{\begin{minipage}{0.37\linewidth}\centering
\textsc{RL-Base}\\Released checkpoint
\end{minipage}} &
\fbox{\begin{minipage}{0.37\linewidth}\centering
\textsc{RL-Indic}\\Separate research run
\end{minipage}}\\[3pt]
Campaign A: \textsc{SFT-Init} vs. \textsc{RL-Base} &
Campaign B: \textsc{RL-Base} vs. \textsc{RL-Indic}
\end{tabular}
\end{tabular}
\caption{Both RL runs start from \textsc{SFT-Init}; \textsc{RL-Indic} is not a continuation of \textsc{RL-Base}. The campaign labels denote evaluation comparisons, not training stages.}
\label{fig:lineage}
\end{figure}

\paragraph{Two comparisons.} Campaign A compares \textsc{SFT-Init} with \textsc{RL-Base} on 34 benchmarks and 55 metrics. Campaign B compares \textsc{RL-Base} with \textsc{RL-Indic}, a separate research run from the same \textsc{SFT-Init} weights. It is not a continuation of \textsc{RL-Base} and is not released. Each campaign uses one harness configuration, model mode, and decoding setup for both models. Figure~\ref{fig:lineage} summarizes the checkpoint lineage and comparisons; Table~\ref{tab:main} keeps the results separate. Accuracy metrics are percentages, and chrF++, chrF, and BLEU use the sacreBLEU 0--100 scale. All score changes are reported in points rather than relative percentages.

\section{What changed after reinforcement learning?}
\label{sec:results}
Table~\ref{tab:main} shows selected results. In campaign A, MATH, a competition-mathematics benchmark~\citep{hendrycks2021math}, improves by 36.07 points and English instruction following on IFEval by 16.45. Several Indic tasks also improve, including translation on FLORES-IN ($+21.78$ chrF++) and translated instruction following ($+12.36$). Math gains are uneven: MATH and MATH-500 improve sharply, while AIME and HMMT move only modestly.

\begin{table}[t]
\centering\footnotesize
\caption{Selected results from two separate comparisons. Campaign A measures the change from \textsc{SFT-Init} to \textsc{RL-Base}. Campaign B measures the change from \textsc{RL-Base} to \textsc{RL-Indic}. Each $\Delta$ is measured within its own campaign, in points. No $\Delta$ reported in this paper is computed across campaigns. Bold highlights the main gains and declines. EM denotes exact match, scored under one reconciled extraction rule across all three checkpoints. Because that rule differs from the one used in earlier records, the text reports passage answering in F1, which is unaffected; Appendix~\ref{app:full} carries both. \mbox{All instruction-following scores use prompt-strict accuracy.}}
\label{tab:main}
\setlength{\tabcolsep}{5pt}
\begin{tabular}{@{}lrrrrrr@{}}\toprule
& \multicolumn{3}{c}{Campaign A} & \multicolumn{3}{c}{Campaign B} \\
\cmidrule(lr){2-4}\cmidrule(l){5-7}
Benchmark & SFT & RL-Base & $\Delta$ & RL-Base & RL-Indic & $\Delta$ \\\midrule
\multicolumn{7}{@{}l}{\textit{English controls}} \\
MATH & 48.69 & 84.76 & $\mathbf{+36.07}$ & 84.76 & 86.76 & $+2.00$ \\
IFEval & 66.36 & 82.81 & $+16.45$ & 82.81 & 86.88 & $+4.07$ \\
XQuAD-English (EM) & 46.97 & 63.70 & $+16.73$ & 63.70 & 61.85 & $-1.85$ \\
\addlinespace[3pt]
\multicolumn{7}{@{}l}{\textit{Indic passage answering}} \\
XQuAD-Indic (EM) & 18.84 & 12.86 & $\mathbf{-5.98}$ & 12.86 & 32.00 & $\mathbf{+19.14}$ \\
IndicQA (EM) & 12.77 & 12.22 & $-0.55$ & 12.22 & 23.46 & $\mathbf{+11.24}$ \\
\addlinespace[3pt]
\multicolumn{7}{@{}l}{\textit{Indic translation and instruction following}} \\
FLORES-IN (chrF++) & 14.55 & 36.33 & $+21.78$ & 36.33 & 39.37 & $+3.04$ \\
IndicIFEval-translated & 37.51 & 49.87 & $+12.36$ & 49.87 & 63.27 & $+13.40$ \\
IndicIFEval-ground & 40.52 & 33.05 & $-7.47$ & 33.05 & 60.00 & $+26.95$ \\
IFEval-Hindi & 52.71 & 50.59 & $-2.12$ & 50.59 & 72.29 & $+21.70$ \\\bottomrule
\end{tabular}
\end{table}

Six campaign-A benchmarks decline, four on Indic QA or instruction following. HumanEval+ is nearly unchanged; BBQ is a bias diagnostic. Figure~\ref{fig:benchmark-deltas} summarizes selected changes from both campaigns; Appendix~\ref{app:full} reports full results.

\section{Passage answering across languages}
\label{sec:boundary}
A general decline in extraction cannot explain the results. English XQuAD improves by 11.71 F1 points while XQuAD-Indic falls by 19.32. Both use the same grading method on corresponding question-answering items. The versions differ in translation, one-shot examples, and language coverage (Appendix~\ref{app:taxonomy}).

The decline is mainly a change of output language. The share of XQuAD-Indic answers written in English rises from 8.7\% to 91.7\%, and a prose-tolerant regrade moves macro F1 by at most 0.24 points, so extraction alone does not account for the loss. IndicQA behaves the same way. XorQA-Indic improves by 27.93 F1 points, but uses Indic questions and answers with English passages.

The instruction-following results also depend on the evaluation data. IndicIFEval has two splits~\citep{jayakumar-etal-2026-indicifeval}. The translated split uses translations of English IFEval prompts. The ground split builds instructions around natively authored Indic text. Neither is passage question answering. The translated split improves by 12.36 points, while the ground split falls by 7.47. These results show why translated and locally authored instructions should be reported separately.

\section{A separate run with Indic training}
\label{sec:indic}
We next tested whether Indic-specific training could improve the tasks that declined. We started \textsc{RL-Indic} from the same \textsc{SFT-Init} checkpoint. We added an Indic instruction-following environment, \texttt{indic\_if}, and language and repetition reward terms. The language term applies only to the answer after the final reasoning block, allowing English reasoning before an Indic answer. English examples target Latin script. Examples with unknown language labels skip the language term. Appendix~\ref{app:rl} gives the coefficients and full configuration.

The mixture also retains a 9.9\% pool of English examples drawn from the four source environments. This was important because an Indic-only dataset with an Indic-only reward had led the model to answer English prompts in Indic script. Retaining English examples was intended to preserve language control on English prompts, rather than reward Indic output regardless of the request.

\begin{figure}[!ht]
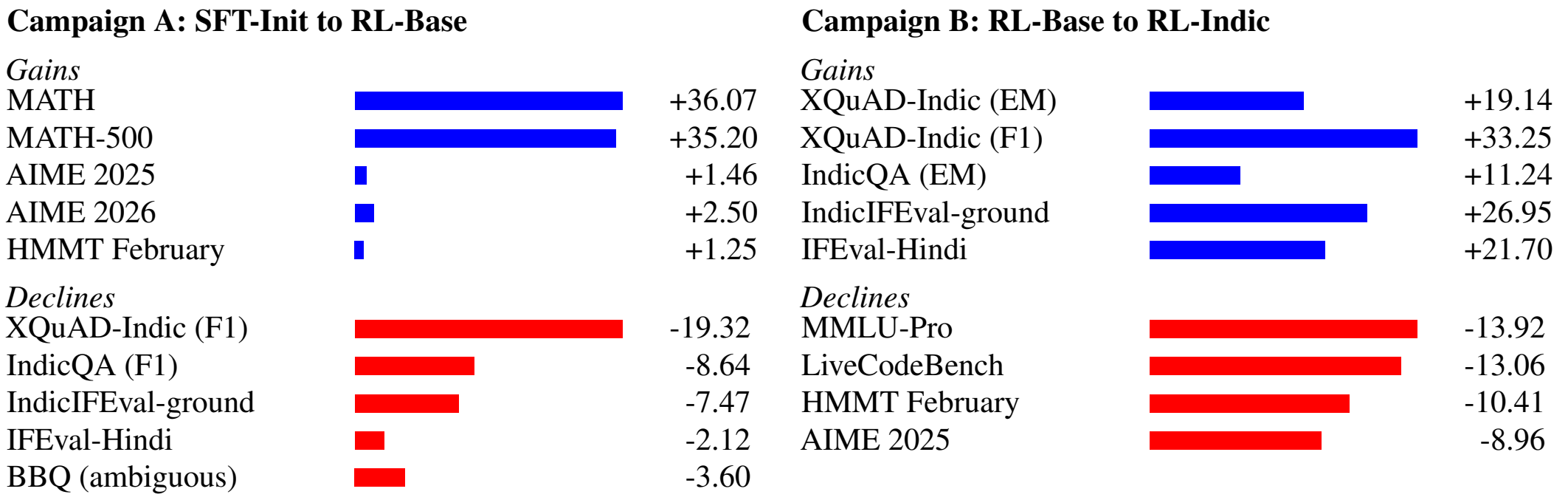

\centering\footnotesize
\begin{minipage}[t]{0.49\linewidth}
\vspace{0pt}
\textbf{Campaign A: SFT-Init to RL-Base}\par\vspace{3pt}
\noindent\textit{Gains}\par
\noindent\begin{tabular}{@{}p{27mm} l r@{}}
MATH & \makebox[24mm][l]{\textcolor{blue}{\rule{24.0mm}{1.3ex}}} & +36.07 \\
MATH-500 & \makebox[24mm][l]{\textcolor{blue}{\rule{23.4mm}{1.3ex}}} & +35.20 \\
AIME 2025 & \makebox[24mm][l]{\textcolor{blue}{\rule{1.0mm}{1.3ex}}} & +1.46 \\
AIME 2026 & \makebox[24mm][l]{\textcolor{blue}{\rule{1.7mm}{1.3ex}}} & +2.50 \\
HMMT February & \makebox[24mm][l]{\textcolor{blue}{\rule{0.8mm}{1.3ex}}} & +1.25
\end{tabular}\par\vspace{3pt}
\noindent\textit{Declines}\par
\noindent\begin{tabular}{@{}p{27mm} l r@{}}
XQuAD-Indic (F1) & \makebox[24mm][l]{\textcolor{red}{\rule{24.0mm}{1.3ex}}} & -19.32 \\
IndicQA (F1) & \makebox[24mm][l]{\textcolor{red}{\rule{10.7mm}{1.3ex}}} & -8.64 \\
IndicIFEval-ground & \makebox[24mm][l]{\textcolor{red}{\rule{9.3mm}{1.3ex}}} & -7.47 \\
IFEval-Hindi & \makebox[24mm][l]{\textcolor{red}{\rule{2.6mm}{1.3ex}}} & -2.12 \\
BBQ (ambiguous) & \makebox[24mm][l]{\textcolor{red}{\rule{4.5mm}{1.3ex}}} & -3.60
\end{tabular}
\end{minipage}\hfill
\begin{minipage}[t]{0.49\linewidth}
\vspace{0pt}
\textbf{Campaign B: RL-Base to RL-Indic}\par\vspace{3pt}
\noindent\textit{Gains}\par
\noindent\begin{tabular}{@{}p{27mm} l r@{}}
XQuAD-Indic (EM) & \makebox[24mm][l]{\textcolor{blue}{\rule{13.8mm}{1.3ex}}} & +19.14 \\
XQuAD-Indic (F1) & \makebox[24mm][l]{\textcolor{blue}{\rule{24.0mm}{1.3ex}}} & +33.25 \\
IndicQA (EM) & \makebox[24mm][l]{\textcolor{blue}{\rule{8.1mm}{1.3ex}}} & +11.24 \\
IndicIFEval-ground & \makebox[24mm][l]{\textcolor{blue}{\rule{19.5mm}{1.3ex}}} & +26.95 \\
IFEval-Hindi & \makebox[24mm][l]{\textcolor{blue}{\rule{15.7mm}{1.3ex}}} & +21.70
\end{tabular}\par\vspace{3pt}
\noindent\textit{Declines}\par
\noindent\begin{tabular}{@{}p{27mm} l r@{}}
MMLU-Pro & \makebox[24mm][l]{\textcolor{red}{\rule{24.0mm}{1.3ex}}} & -13.92 \\
LiveCodeBench & \makebox[24mm][l]{\textcolor{red}{\rule{22.5mm}{1.3ex}}} & -13.06 \\
HMMT February & \makebox[24mm][l]{\textcolor{red}{\rule{17.9mm}{1.3ex}}} & -10.41 \\
AIME 2025 & \makebox[24mm][l]{\textcolor{red}{\rule{15.4mm}{1.3ex}}} & -8.96
\end{tabular}
\end{minipage}
\caption{Selected benchmark changes in points. Bars are proportional within each gains or declines panel; scales differ between panels, and metrics are benchmark-native. Deltas are computed within each campaign only.}
\label{fig:benchmark-deltas}
\end{figure}

In campaign B, \textsc{RL-Indic} improves all four Indic benchmarks that declined in campaign A, as well as MATH, but loses ground on MMLU-Pro, LiveCodeBench, HMMT February, and AIME 2025. Figure~\ref{fig:benchmark-deltas} shows the selected point changes for both campaigns.

Comparing \textsc{RL-Indic} with \textsc{SFT-Init} requires the two campaigns to be comparable (Appendix~\ref{app:head}). Several training choices changed together, so no reward term is isolated.

\section{Script compliance and answer formatting}
\label{sec:convention}
Script compliance means using the requested writing system in the final answer. Better task scores did not guarantee better script compliance. On translated IndicIFEval, in-script output is 95.1\% for \textsc{SFT-Init} and 84.6\% for \textsc{RL-Indic}, despite the added language reward. \textsc{RL-Base}'s ground-split strict accuracy ranges from 28.66 to 63.88 across three non-matched campaigns; the spread is unresolved and should not be read as a model change (Table~\ref{tab:script-campaigns}).

Early \textsc{RL-Base} rollouts offer one reason to examine script choice. Within 1,148 prompt groups, Latin-script answers receive 0.168 more reward on average than Indic-script answers (95\% CI $[0.149, 0.187]$). Structured-output and workplace-assistant environments account for 95\% of this pooled difference. Their graders can accept outputs such as numerals or API names without Indic text. The paired generations differ, and reward scales vary by environment (Appendix~\ref{app:differential}).

Answer formatting can also affect benchmark scores. A boxed answer marks the final mathematical result with the LaTeX command \texttt{\textbackslash boxed\{...\}}. Across 5,000 MATH problems, overall accuracy rises from 51.00 to 85.80, yet accuracy among answers with a boxed final result stays near 91--93\%. This does not establish the correctness of unboxed answers. Regrading all outputs would help distinguish formatting changes from changes in ability. For Indic passage answering that check has been run, and it points to output language.

\section{Limitations and practical implications}
\label{sec:limits}
\label{sec:deploy}
This is a single-seed case study, not an ablation of individual reward terms. We do not report human evaluation, contamination audits, per-item confidence intervals, or per-language breakdowns. The reward differential covers early rollouts rather than the full run. We omit BharatMath, which is pending a harness revision.

For Indic document assistance, gains in math, translation, or instruction following are not enough. Evaluation should include at least one passage-answering task per target language, using the checkpoint and evaluation configuration that will be reported. Native-script and romanized inputs should be tested separately, as should translated and locally authored prompts. Report script compliance and answer-format rates alongside accuracy. In this study, added Indic training improved passage-answering scores but did not ensure reliable script compliance.

\typeout{MAIN-END-PAGE=\thepage}

\clearpage
\appendix
\section{Checkpoints and scope}
\label{app:scope}

\textsc{SFT-Init} is the supervised checkpoint used to initialize both RL runs. \textsc{RL-Base} is the released GRPO checkpoint at step 40, trained on four environments. \textsc{RL-Indic} is a separate research run from the same \textsc{SFT-Init} weights. It is not a continuation of \textsc{RL-Base} and has not been released.

\section{Supervised initialization}
\label{app:sft}

\begin{table}[!ht]
\centering\small
\caption{Training record for \textsc{SFT-Init}, the supervised checkpoint used to initialize both RL runs.}
\begin{tabular}{ll}\toprule
Setting & Value \\\midrule
Export point & partway through the final pass, verified from export metadata \\
Scheduled sequence positions & 80.53B ($2{,}400\times256\times131{,}072$), including padding \\
Approximate loss-bearing positions & $\sim$15B (18.2\%, assistant-only mask) \\
Trainable-rate measurement & 95,515 / 524,288 positions over four packed rows \\
Intra-pack packing efficiency & 99.98\% \\
Prepared packed capacity & 92.12B positions across 702,824 sequences \\
Packed rows / scheduled documents & 702,824 / 15,160,245 \\
Configured iters / endpoint / RL export & 2,752 / 2,665 / 2,400 \\
Epoch traversal & sequential bucket blocks; no repeat or wrap-around \\
Sequence length / precision & 131,072 / BF16 \\
GPUs / parallelism & 256 H200; TP=1, PP=1, CP=4, EP=8, DP=64 \\
Global / micro batch & 256 / 1 \\
Learning-rate schedule & cosine, $5\times10^{-7}$ peak to $1\times10^{-7}$ minimum \\
Optimizer & Adam, $\beta_2=0.95$, $\epsilon=10^{-8}$, weight decay 0.1, clip 1.0 \\
Data organization & 15 broadly homogeneous curriculum buckets \\
Materialized rows / language balance & 15,214,121; 64.0\% English, 36.0\% Indic \\
Loss at RL-init export & 0.613583 (per-step); median 0.606070 over iters 2,350--2,450 \\
Retained SFT curriculum compute & $\sim$17,205 GPU-hours (entire curriculum, not only to iter 2,400) \\
\bottomrule
\end{tabular}
\end{table}

\paragraph{Iteration count.} Without bucket boundaries, the iteration count would be $\lceil 702{,}824/256 \rceil = 2{,}746$. The configured count is 2,752 because the 15 curriculum buckets are processed separately. Each bucket ends at a step boundary, giving $\sum_b \lceil N_b/256 \rceil$ steps in total. Its final partial batch is padded with EOS sequences whose loss mask is zero. The extra slots are padding, not repeated data or the result of \texttt{drop\_last}. This padding occupies about 0.2\% of the scheduled epoch's batch slots.

\paragraph{Loss-bearing positions.} The illustrative estimate of roughly 15B loss-bearing positions uses an 18.2\% rate measured on four packed parquet rows before runtime padding.

\paragraph{Training corpus.} The data cover English and eight Indic languages: Bengali, Gujarati, Hindi, Kannada, Malayalam, Marathi, Tamil, and Telugu. Indic examples use native and romanized script. Tasks include question answering, reasoning, mathematics, summarization, instruction following, tool interaction, and multi-turn dialogue. Hindi accounts for 11.1\% of rows, and Tamil, Telugu, Kannada, and Malayalam each account for 4.8\%. These percentages describe rows, not token or loss shares.

User-facing prose is translated, but code, mathematical spans, tool schemas, function names, and machine-readable arguments are preserved byte for byte. Structurally incomplete translations are set aside rather than replaced with English. Training loss applies only to the selected assistant spans. System and user messages remain available as context but are masked from the loss. Tool chains remain intact so that intermediate actions and observations are learned together.

\paragraph{Script compliance before RL.} At initialization, in-script output rates range from 93.5\% to 99.1\% in one multilingual harness. A later matched IndicIFEval-ground evaluation records 99.9\%. These measurements establish the script behavior before RL. As discussed in \S\ref{sec:convention}, \textsc{RL-Base} has no explicit reward term to preserve that behavior during training.

\section{Passage language and prompt source}
\label{app:taxonomy}

\paragraph{Indic passage answering.} XQuAD-Indic and IndicQA provide a passage in the target language. The grader compares the generated answer with the reference using exact match or token F1. Both benchmarks decline in campaign A. Section~\ref{sec:boundary} examines these passage-answering declines.

XQuAD-Indic is XQuAD-IN from IndicGenBench~\citep{singh-etal-2024-indicgenbench}. It uses professional human translations of XQuAD's 1{,}190 SQuAD-v1.1-dev items into twelve Indic languages. Its Hindi split is translated independently of the split in \citet{artetxe-etal-2020-cross}, so the scores are not directly comparable with published XQuAD-hi results. Reference answers are translated strings without character offsets. Scoring therefore uses string-level exact match and F1 rather than passage-span extraction.

Four of the benchmark's twelve Indic languages fall outside the model's eight-language training coverage. The English control uses the original XQuAD over the same 1{,}190 items. Its one-shot example comes from the evaluation split, while the Indic version draws from a separate pool. These differences limit how closely the English and Indic results isolate the effect of language.

\paragraph{Cross-lingual passage answering.} XorQA-Indic uses English passages from the same suite. Only the question and answer are Indic. This gain does not establish better use of Indic passages.

\paragraph{Instruction following.} IndicIFEval~\citep{jayakumar-etal-2026-indicifeval} covers fourteen Indic languages in two splits. The translated split translates English IFEval prompts. The ground split starts with natively authored Indic documents that satisfy a verifiable property. A model then writes an instruction around that property, using native words and structures rather than translated constraints. Neither split supplies a passage against which the answer is graded. The distinction is prompt source, not passage answering. We do not assign IFEval-Hindi to either prompt-source category.

\clearpage
\section{Extended matched evaluation}
\label{app:full}

{\footnotesize\setlength{\tabcolsep}{4pt}
\begin{longtable}{llrrr}
\caption{Campaign A: \textsc{SFT-Init} and the released \textsc{RL-Base} checkpoint, evaluated with the same harness, model mode, and decoding configuration. HumanEval+ is omitted. Each IndicIFEval-ground row combines identical prompt-level and instruction-level scores. Changes are RL minus SFT. Accuracy metrics are percentages. The chrF++, chrF, and BLEU values are sacreBLEU scores on a 0--100 scale, not percentages. For rows marked $\downarrow$, lower scores are better, so a negative change indicates improvement. \mbox{Bold marks the better score in each row.}}\label{tab:full}\\
\toprule
Benchmark & Metric & Final SFT & RL ckpt & $\Delta$ (RL $-$ SFT) \\
          &        & ckpt      & step 40 & \\
\midrule
\endfirsthead
\toprule Benchmark & Metric & Final SFT & RL ckpt & $\Delta$ (RL $-$ SFT) \\\midrule
\endhead
\addlinespace[2pt]
\multicolumn{5}{l}{\textit{Math \& Reasoning}} \\
MATH & exact match & 48.69 & \textbf{84.76} & $+36.07$ \\
MATH-500 & exact match & 48.80 & \textbf{84.00} & $+35.20$ \\
GSM8K & exact match & 85.75 & \textbf{94.31} & $+8.56$ \\
MGSM-English & exact match & 91.60 & \textbf{98.40} & $+6.80$ \\
MGSM-Indic & exact match & 79.60 & \textbf{84.60} & $+5.00$ \\
AIME 2025 & pass@1 & 72.92 & \textbf{74.38} & $+1.46$ \\
AIME 2026 & pass@1 & 72.08 & \textbf{74.58} & $+2.50$ \\
Beyond AIME & pass@1 & 48.50 & \textbf{53.56} & $+5.06$ \\
HMMT February 2025 & pass@1 & 59.58 & \textbf{60.83} & $+1.25$ \\
HMMT November 2025 & pass@1 & 58.54 & \textbf{63.33} & $+4.79$ \\
GPQA Diamond & pass@1 & 38.19 & \textbf{65.03} & $+26.84$ \\
\addlinespace[2pt]
\addlinespace[2pt]
\multicolumn{5}{l}{\textit{Code}} \\
LiveCodeBench & pass@1 & 51.35 & \textbf{61.27} & $+9.92$ \\
MBPP+ & pass@1 & 75.24 & \textbf{78.78} & $+3.54$ \\
\addlinespace[2pt]
\addlinespace[2pt]
\multicolumn{5}{l}{\textit{Knowledge and MCQ}} \\
MMLU$^{\dagger}$ & exact match & 77.21 & \textbf{86.68} & $+9.47$ \\
MMLU-Pro & accuracy & 66.67 & \textbf{77.26} & $+10.59$ \\
IndicMMLU-Pro & accuracy & 47.07 & \textbf{68.88} & $+21.81$ \\
MILU & accuracy & 73.16 & \textbf{78.33} & $+5.17$ \\
ARC-Challenge & exact match & 93.26 & \textbf{95.90} & $+2.64$ \\
AGIEval English & exact match & 60.02 & \textbf{73.47} & $+13.45$ \\
BhashaBench-Multi & accuracy & 42.06 & \textbf{49.22} & $+7.16$ \\
\addlinespace[2pt]
\addlinespace[2pt]
\multicolumn{5}{l}{\textit{Instruction Following}} \\
IFEval & prompt strict & 66.36 & \textbf{82.81} & $+16.45$ \\
IFEval & prompt loose & 67.65 & \textbf{83.73} & $+16.08$ \\
IFEval & inst.\ strict & 73.26 & \textbf{87.17} & $+13.91$ \\
IFEval & inst.\ loose & 74.22 & \textbf{87.77} & $+13.55$ \\
IFEval-Hindi & prompt strict & \textbf{52.71} & 50.59 & $-2.12$ \\
IFEval-Hindi & prompt loose & \textbf{53.42} & 51.30 & $-2.12$ \\
IFEval-Hindi & inst.\ strict & \textbf{60.51} & 59.64 & $-0.87$ \\
IFEval-Hindi & inst.\ loose & \textbf{61.62} & 60.27 & $-1.35$ \\
IndicIFEval-translated & prompt strict & 37.51 & \textbf{49.87} & $+12.36$ \\
IndicIFEval-translated & prompt loose & 38.88 & \textbf{50.87} & $+11.99$ \\
IndicIFEval-translated & inst.\ strict & 47.68 & \textbf{61.65} & $+13.97$ \\
IndicIFEval-translated & inst.\ loose & 49.04 & \textbf{62.43} & $+13.39$ \\
IndicIFEval-ground & prompt/inst.\ strict & \textbf{40.52} & 33.05 & $-7.47$ \\
IndicIFEval-ground & prompt/inst.\ loose & \textbf{43.59} & 33.43 & $-10.16$ \\
IFBench & prompt strict & 21.67 & \textbf{44.33} & $+22.66$ \\
IndicParam & accuracy & 32.94 & \textbf{39.28} & $+6.34$ \\
\addlinespace[2pt]
\addlinespace[2pt]
\multicolumn{5}{l}{\textit{QA and Translation}} \\
XQuAD-English & exact match & 46.97 & \textbf{63.70} & $+16.73$ \\
XQuAD-English & F1 & 67.15 & \textbf{78.86} & $+11.71$ \\
XQuAD-Indic & exact match & \textbf{18.84} & 12.86 & $-5.98$ \\
XQuAD-Indic & F1 & \textbf{36.51} & 17.19 & $-19.32$ \\
XorQA-Indic & exact match & 6.67 & \textbf{35.24} & $+28.57$ \\
XorQA-Indic & F1 & 16.16 & \textbf{44.09} & $+27.93$ \\
IndicQA & exact match & \textbf{12.77} & 12.22 & $-0.55$ \\
IndicQA & F1 & \textbf{26.36} & 17.72 & $-8.64$ \\
FLORES-IN & chrF++ & 14.55 & \textbf{36.33} & $+21.78$ \\
FLORES-IN & BLEU & 3.64 & \textbf{14.94} & $+11.30$ \\
CrossSum-IN & chrF & 7.00 & \textbf{12.81} & $+5.81$ \\
\addlinespace[2pt]
\addlinespace[2pt]
\multicolumn{5}{l}{\textit{Bias diagnostic}} \\
BBQ & accuracy (disamb.) & 94.13 & \textbf{97.70} & $+3.57$ \\
BBQ & accuracy (amb.) & \textbf{92.29} & 88.69 & $-3.60$ \\
BBQ & bias score (disamb.)$\downarrow$ & 0.99 & \textbf{0.60} & $-0.39$ \\
BBQ & bias score (amb.)$\downarrow$ & \textbf{4.33} & 7.39 & $+3.06$ \\
\addlinespace[2pt]
\bottomrule
\end{longtable}}

{\footnotesize $^{\dagger}$MMLU uses generative exact match. Its scores are not directly comparable with loglikelihood-based accuracy.\par}

\section{Reinforcement-learning configurations}
\label{app:rl}

\begin{table}[!ht]
\centering\small
\caption{Released \textsc{RL-Base} configuration. Each environment returns its own binary or task-specific reward. \mbox{No additional reward shaping is applied.}}
\begin{tabular}{@{}lp{8.6cm}@{}}\toprule
Setting & Value \\\midrule
Initialization & \textsc{SFT-Init} \\
Algorithm & GRPO; leave-one-out baseline, per-group Bessel-corrected standard-deviation normalization, divisor $\sigma+10^{-6}$ \\
Resolved estimator flags & \texttt{use\_leave\_one\_out\_baseline=true}; \texttt{normalize\_rewards=true} \\
Generations per prompt & 16 \\
Dynamic sampling & enabled; zero-variance groups resampled \\
Sampling temperature & 1.0 \\
Environments & judged mathematics; structured outputs; workplace assistant; MCQA \\
Language / length / repetition reward & none \\
Inference placement & colocated generation and training workers \\
Recorded compute & 763 GPU-hours \\
Validation reward & 0.508 at step 10 $\rightarrow$ 0.719 at step 40 (not accuracies) \\
\bottomrule
\end{tabular}
\end{table}

\begin{table}[!ht]
\centering\small
\caption{\textsc{RL-Indic} configuration and input mixture. The rows are filtered by difficulty and then capped within each environment. Shares describe the resulting pool, not preset mixture weights. Rounded shares sum to 100.1\%, while the row counts sum to exactly 15,300.}
\begin{tabular}{@{}lp{4.3cm}@{\ }lrr@{}}\toprule
Setting & Value & Pool & Rows & Share \\\midrule
Added environment & \texttt{indic\_if} & Mathematics & 6,192 & 40.5\% \\
Reward & $R_{\text{task}}\!+\!0.3R_{\text{lang}}\!-\!0.0R_{\text{len}}\!-\!0.3R_{\text{rep}}$ & Indic instruction following & 3,179 & 20.8\% \\
Malformed / abandoned & $-0.25$ & Structured outputs & 1,932 & 12.6\% \\
Advantage std.\ floor & 0.05 & Workplace assistant & 1,804 & 11.8\% \\
Steps / rows / validation & 40 / 15,300 / 256 & English guardrail & 1,509 & 9.9\% \\
Recorded rollouts & 251,906 & MCQA & 684 & 4.5\% \\
Hardware / wall time & 16$\times$8 H200 / 5h34m & \textbf{Total} & \textbf{15,300} & \textbf{100\%} \\
Max generated tokens & 16,384 & & & \\
Max RL sequence length & 49,152 (rollout cap only) & & & \\
Checkpoint cadence & every 5 steps & & & \\
Validation reward & 0.320 $\rightarrow$ 0.649 at step 40 & & & \\
Checkpoint selection & held-out validation set & & & \\
\bottomrule
\end{tabular}
\end{table}

The RL sequence cap of 49,152 tokens limits rollout collection. It does not change the model's trained or exported context length of 131,072 tokens.

$R_{\text{task}}$ is the fraction of constraints satisfied. $R_{\text{lang}}$ is the share of answer words in the target script, measured only after the final reasoning block. A language-identification check distinguishes languages that share a script. Examples with unknown language labels skip this term. $R_{\text{rep}}$ detects repeated $n$-gram bursts and character runs across the whole generation.

The checkpoint-selection rule preferred step 30 unless step 40 retained MATH performance. Step 40 met that condition and also improved GSM8K, IFEval, and both IndicIFEval splits. We therefore report step 40 for the comparison with the released checkpoint.

These tables report the input mixture. Dynamic sampling discards groups in which every completion passes or every completion fails, so the training proportions can differ from the listed shares.

\section{Reward differences between Latin- and Indic-script answers}
\label{app:differential}

We compare rewards for Latin- and Indic-script answers to the same prompts, using early-training rollouts from one seed. We report each environment separately because reward scales differ.

\subsection{Estimator}

We classify the script of the answer after the final reasoning block and retain prompt groups with both Latin- and Indic-script completions. English reasoning followed by an Indic answer counts as an Indic-script completion. The paired difference and its average are
\begin{equation}
d_g \;=\; \frac{1}{|E_g|}\sum_{i \in E_g} r_i \;-\; \frac{1}{|I_g|}\sum_{i \in I_g} r_i ,
\qquad
\bar{d} \;=\; \frac{1}{|G^{*}|}\sum_{g \in G^{*}} d_g ,
\end{equation}
Here $E_g$ and $I_g$ are the Latin- and Indic-script completions in group $g$. $G^{*}$ contains the groups with both scripts, and $r_i$ is the reward for completion $i$. The environment breakdown is $\bar d = \sum_e w_e \bar d_e$, where $w_e$ is the environment's share of paired groups and $\bar d_e$ is its mean reward difference.

Pairing holds the prompt fixed, but the completions can differ in content as well as script. The estimate therefore measures an association, not the effect of changing script alone.

\subsection{Results}

\begin{table}[!ht]
\centering\small
\caption{Paired reward differences in the retained rollout sample. Each environment contributes its mean difference weighted by its share of paired prompt groups. Here $w_e$ is that share within these \textsc{RL-Base} rollouts, not the environment's share of training rows in the \textsc{RL-Indic} input mixture. Uncertainty is reported as a standard error (SE) and a confidence interval (CI).}
\begin{tabular}{@{}lrrr@{}}\toprule
Environment & $w_e$ & $\bar d_e$ & contribution $w_e\bar d_e$ \\\midrule
Structured outputs & 43.4\% & $+0.2588$ & $+0.1123$ \\
Workplace assistant & 19.3\% & $+0.2483$ & $+0.0480$ \\
MCQA & 7.5\% & $+0.1030$ & $+0.0077$ \\
Mathematics with judge & 29.8\% & $+0.0006$ & $+0.0002$ \\\midrule
\textbf{Aggregate $\bar d$} & & \multicolumn{2}{r}{$\mathbf{+0.1681}$ \; (SE 0.0095)} \\
95\% bootstrap CI & & \multicolumn{2}{r}{$[+0.1489, +0.1871]$} \\
Paired groups $|G^{*}|$ & & \multicolumn{2}{r}{1,148} \\
Unpaired contrast & & \multicolumn{2}{r}{$+0.1373$} \\\bottomrule
\end{tabular}
\end{table}

Structured outputs and the workplace assistant account for 95.4\% of the pooled difference. The contribution shares depend on the environments' reward scales. GRPO uses normalized advantages, so these raw-reward differences do not directly measure the resulting training pressure.

\paragraph{Answer-length threshold.} The main estimate requires at least 20 script-bearing characters and gives $\bar d = +0.1681$. Excluding rather than including ambiguous spans changes it by 0.0009. Including all spans reduces the estimate to $+0.066$ and makes the mathematics difference negative ($-0.082$). Short outputs such as numerals and option letters often reflect format rather than language choice, so the reported estimate applies to the filtered set of completions.

\section{Comparing the RL runs and tool-use results}
\label{app:head}

Table~\ref{tab:threeway} reports the 34 benchmarks shared by both campaigns. Across the 33 shared metrics on a 0--1 scale, the arithmetic mean is 0.661233 for \textsc{RL-Indic} and 0.661188 for \textsc{RL-Base}, a difference of 0.000045. These nearly equal averages hide substantial differences. \textsc{RL-Indic} scores higher on instruction following, translation, and several Indic generation and QA tasks. \textsc{RL-Base} remains stronger on MMLU-Pro, LiveCodeBench, HMMT, AIME, GPQA-Diamond, MMLU, and MBPP+.

\paragraph{Comparisons across campaigns.} Campaign A measures $\textsc{RL-Base}-\textsc{SFT-Init}$, and campaign B measures $\textsc{RL-Indic}-\textsc{RL-Base}$. The reconstructed comparison of \textsc{RL-Indic} with \textsc{SFT-Init} assumes the campaigns are comparable. The shared \textsc{RL-Base} scores match on all 34 benchmarks.

\begin{table}[p]
\centering\footnotesize
\caption{Three-way matched comparison over the 34 benchmarks scored in both campaigns. The
\textsc{RL-Base} column is identical, value for value, to the corresponding column of
Table~\ref{tab:full}; that agreement across all 34 is our evidence that the two campaigns share an
evaluation state, and is corroboration rather than a provenance record. Metrics are benchmark-native and are the single reported metric per
benchmark where several exist; the complete 55-metric \textsc{SFT-Init}/\textsc{RL-Base} pair is
Table~\ref{tab:full}.}
\label{tab:threeway}
\setlength{\tabcolsep}{4pt}
\begin{tabular}{@{}lrrrrr@{}}\toprule
Benchmark & SFT & RL-Base & $\Delta_{\text{Base}}$ & RL-Indic & $\Delta_{\text{Indic}}$ \\\midrule
\addlinespace[2pt]\multicolumn{6}{@{}l}{\textit{Math \& Reasoning}} \\
MATH & 48.69 & 84.76 & $+36.07$ & 86.76 & $+38.07$ \\
MATH-500 & 48.80 & 84.00 & $+35.20$ & 87.60 & $+38.80$ \\
GSM8K & 85.75 & 94.31 & $+8.56$ & 91.58 & $+5.83$ \\
MGSM-English & 91.60 & 98.40 & $+6.80$ & 97.60 & $+6.00$ \\
MGSM-Indic & 79.60 & 84.60 & $+5.00$ & 86.40 & $+6.80$ \\
AIME 2025 & 72.92 & 74.38 & $+1.46$ & 65.42 & $-7.50$ \\
AIME 2026 & 72.08 & 74.58 & $+2.50$ & 67.50 & $-4.58$ \\
Beyond AIME & 48.50 & 53.56 & $+5.06$ & 46.37 & $-2.13$ \\
HMMT February 2025 & 59.58 & 60.83 & $+1.25$ & 50.42 & $-9.16$ \\
HMMT November 2025 & 58.54 & 63.33 & $+4.79$ & 58.13 & $-0.41$ \\
GPQA Diamond & 38.19 & 65.03 & $+26.84$ & 60.04 & $+21.85$ \\
\addlinespace[2pt]\multicolumn{6}{@{}l}{\textit{Code}} \\
LiveCodeBench & 51.35 & 61.27 & $+9.92$ & 48.21 & $-3.14$ \\
MBPP+ & 75.24 & 78.78 & $+3.54$ & 74.29 & $-0.95$ \\
HumanEval+ & 88.66 & 88.54 & $-0.12$ & 88.29 & $-0.37$ \\
\addlinespace[2pt]\multicolumn{6}{@{}l}{\textit{Knowledge and MCQ}} \\
MMLU$^{\dagger}$ & 77.21 & 86.68 & $+9.47$ & 81.67 & $+4.46$ \\
MMLU-Pro & 66.67 & 77.26 & $+10.59$ & 63.34 & $-3.33$ \\
IndicMMLU-Pro & 47.07 & 68.88 & $+21.81$ & 61.91 & $+14.84$ \\
MILU & 73.16 & 78.33 & $+5.17$ & 76.90 & $+3.74$ \\
ARC-Challenge & 93.26 & 95.90 & $+2.64$ & 94.97 & $+1.71$ \\
AGIEval English & 60.02 & 73.47 & $+13.45$ & 70.52 & $+10.50$ \\
BhashaBench-Multi & 42.06 & 49.22 & $+7.16$ & 48.33 & $+6.27$ \\
\addlinespace[2pt]\multicolumn{6}{@{}l}{\textit{Instruction following}} \\
IFEval (prompt strict) & 66.36 & 82.81 & $+16.45$ & 86.88 & $+20.52$ \\
IFEval-Hindi (prompt strict) & 52.71 & 50.59 & $-2.12$ & 72.29 & $+19.58$ \\
IndicIFEval-translated (p.s.) & 37.51 & 49.87 & $+12.36$ & 63.27 & $+25.76$ \\
IndicIFEval-ground (p.s.) & 40.52 & 33.05 & $-7.47$ & 60.00 & $+19.48$ \\
IFBench (prompt strict) & 21.67 & 44.33 & $+22.66$ & 50.33 & $+28.66$ \\
IndicParam & 32.94 & 39.28 & $+6.34$ & 38.09 & $+5.15$ \\
\addlinespace[2pt]\multicolumn{6}{@{}l}{\textit{QA and translation}} \\
XQuAD-English (EM) & 46.97 & 63.70 & $+16.73$ & 61.85 & $+14.88$ \\
XQuAD-Indic (EM) & 18.84 & 12.86 & $-5.98$ & 32.00 & $+13.16$ \\
XorQA-Indic (EM) & 6.67 & 35.24 & $+28.57$ & 27.86 & $+21.19$ \\
IndicQA (EM) & 12.77 & 12.22 & $-0.55$ & 23.46 & $+10.69$ \\
FLORES-IN (chrF++) & 14.55 & 36.33 & $+21.78$ & 39.37 & $+24.82$ \\
CrossSum-IN (chrF) & 7.00 & 12.81 & $+5.81$ & 15.06 & $+8.06$ \\
\addlinespace[2pt]\multicolumn{6}{@{}l}{\textit{Bias diagnostic}} \\
BBQ (accuracy, amb.) & 92.29 & 88.69 & $-3.60$ & 88.59 & $-3.70$ \\
\bottomrule
\end{tabular}

\footnotesize $^{\dagger}$Generative exact match.
\end{table}

\begin{table}[!ht]
\centering\small
\caption{Tool-use and multi-turn benchmarks for \textsc{RL-Base} and \textsc{RL-Indic}. Scores are on a 0--1 scale. Changes are percentage points, with positive values favoring \textsc{RL-Indic}. \mbox{Bold marks the higher score in each row.}}
\begin{tabular}{lrrr}\toprule
Benchmark & \textsc{RL-Base} & \textsc{RL-Indic} & $\Delta$ \\\midrule
MultiChallenge & 0.4925 & \textbf{0.5000} & $+0.75$ \\
TauBench airline & 0.3733 & \textbf{0.4400} & $+6.67$ \\
TauBench retail & 0.1520 & \textbf{0.2807} & $+12.87$ \\
TauBench telecom & 0.0936 & \textbf{0.3626} & $+26.90$ \\
SWE-bench Verified & \textbf{0.2280} & 0.1220 & $-10.60$ \\\bottomrule
\end{tabular}
\end{table}

\begin{table}[!ht]
\centering\small
\caption{BFCL v4 scores for \textsc{SFT-Init} and \textsc{RL-Indic}. The overall score falls by 2.65 points, with the largest losses in categories requiring parallel tool calls.}
\begin{tabular}{lrr}\toprule
Category & \textsc{SFT-Init} & \textsc{RL-Indic} \\\midrule
Headline, renormalized over 17 categories & 53.81 & 51.16 \\
Parallel / parallel-multiple & 69.5 / 80.0 & 0.5 / 0.5 \\
Live parallel / live parallel-multiple & 75.0 / 75.0 & 0.0 / 0.0 \\
Simple Python / multiple & 86.8 / 81.0 & 94.5 / 93.0 \\
Multi-turn base & 34.0 & 58.0 \\\bottomrule
\end{tabular}
\end{table}

In the BFCL parallel-call cases, the model emits one tool call where the benchmark expects two in the same assistant message. The syntax is valid. Given a second turn, it issues the missing call in 95 of 96 examined cases. This suggests a change in output convention, consistent with the training environment's one-action-per-turn format, rather than a general loss of tool-use ability.

For requests that could be handled with parallel calls, this behavior adds an extra interaction round. Training examples that require several independent calls in one response would test whether the output convention can be changed without losing the gains on sequential tool use.

\section{Script compliance across campaigns}
\label{app:script}

\begin{table}[!ht]
\centering\small
\caption{\textsc{RL-Base} scores across three IndicIFEval evaluation campaigns. Dashes mark measurements not retained in the available record. The dedicated campaign records 65.62 under the loose criterion. The paired quantization campaign supplies the campaign-A IndicIFEval-ground scores in Table~\ref{tab:main}; the other two rows enter no reported delta. The differences between campaigns remain unresolved, so comparisons require a matched evaluation setup.}
\label{tab:script-campaigns}
\begin{tabular}{lrrr}\toprule
Campaign & Ground strict & Translated strict & Ground in-script \\\midrule
Initial released-run matched campaign & 28.66 & 47.45 & 5.4\% \\
Dedicated IndicIFEval campaign & 63.88 & --- & 54.2\% \\
Paired quantization campaign, BF16 baseline & 33.05 & 49.87 & --- \\\bottomrule
\end{tabular}
\end{table}

Adding Indic training did not reliably preserve script compliance. From \textsc{SFT-Init} to \textsc{RL-Indic}, the native-Indic in-script rate falls from 99.9\% to 98.7\%. On translated IndicIFEval, it falls from 95.1\% to 84.6\%. These comparisons use \textsc{SFT-Init}, not \textsc{RL-Base}, as the baseline.

On the translated split, the reconstructed prompt-strict accuracy gain over \textsc{SFT-Init} is 25.76 points. It depends on the cross-campaign assumption in Appendix~\ref{app:head}. The directly measured gain over \textsc{RL-Base} is 13.40 points. Higher task accuracy does not establish better language or script control.

\end{document}